**Using Machine Learning to Develop Smart Reflex Testing Protocols**

Matthew McDermott[1,2+], Anand Dighe[3,4,5], Peter Szolovits[1], Yuan Luo[6]*, Jason Baron[3,4]*

1. MIT Computer Science and Artificial Intelligence Lab, Boston, Massachusetts, USA.
2. Harvard Medical School, Department of Biomedical Informatics.
3. Department of Pathology, Massachusetts General Hospital, Boston, Massachusetts, USA.
4. Harvard Medical School, Boston, Massachusetts, USA.
5. MGB HealthCare System, Somerville, Massachusetts, USA.
6. Department of Preventive Medicine, Northwestern University, Chicago, IL, USA.

*Drs. Baron and Luo are co-corresponding authors

+ Dr. McDermott was associated with the MIT affiliation during the time the majority of the work was performed but had switched to the Harvard Medical School affiliation by the time of manuscript revision and submission.

Address reprint requests to**:**

Jason Baron, MD

Massachusetts General Hospital

55 Fruit Street

Boston, MA 02114

(314) 276-0944

jmbaron@partners.org

**Key Words:** Machine learning, Clinical decision support, Laboratory test ordering, Imputation, Missing data, Ferritin, Computational Pathology

**Word Count (main text):** xx

**Tables:** 4

**Figures:** 5

**Supplemental Figures:** 1

**ABSTRACT**

**Objective**: Reflex testing protocols allow clinical laboratories to perform second line diagnostic tests on existing specimens based on the results of initially ordered tests. Reflex testing can support optimal clinical laboratory test ordering and diagnosis. In current clinical practice, reflex testing typically relies on simple "if-then" rules; however, this limits their scope since most test ordering decisions involve more complexity than a simple rule will allow. Here, using the analyte ferritin as an example, we propose an alternative machine learning-based approach to "smart" reflex testing with a wider scope and greater impact than traditional rule-based approaches.

**Methods:** Using patient data, we developed a machine learning model to predict whether a patient getting CBC testing will also have ferritin testing ordered, consider applications of this model to "smart" reflex testing, and evaluate the model by comparing its performance to possible rule-based approaches.

**Results:** Our underlying machine learning models performed moderately well in predicting ferritin test ordering and demonstrated greater suitability to reflex testing than rule-based approaches. Using chart review, we demonstrate that our model may improve ferritin test ordering. Finally, as a secondary goal, we demonstrate that ferritin test results are missing not at random (MNAR), a finding with implications for unbiased imputation of missing test results.

**Conclusions:** Machine learning may provide a foundation for new types of reflex testing with enhanced benefits for clinical diagnosis and laboratory utilization management.

1. INTRODUCTION

Appropriate clinical laboratory test selection is critical to high-quality clinical care. [1] Since many, if not most, clinical decisions rely in significant part on clinical laboratory data, a physician's failure to order necessary tests can lead to delayed, overlooked, or incorrect diagnoses; unnecessary downstream testing; and improper or foregone treatments. [1] For example, a failure to order iron studies in patients with findings suggestive of iron deficiency may lead to an overlooked diagnosis of iron deficiency and delayed treatment. This may in turn subject the patient to unnecessary symptoms and potentially even significant morbidity and mortality. Ordering unnecessary tests is also problematic. [2] Test over-utilization not only wastes physician, nursing, and laboratory resources, potentially subjecting patients and the healthcare system to financial harm, but it can also lead to false positive results, unnecessary patient anxiety, unnecessary downstream workups, and in some cases even inappropriate treatments. In addition, physicians may overlook important test results among unneeded ones in patients with excessive testing data. [3 4] To improve test utilization and patient care, clinical laboratories and health systems can pursue strategies including the development of institutional guidelines and best practices, the delivery of focused clinician education, the deployment of clinical decision support and the offering of "reflex testing" protocols. [1 3 5]

Reflex testing enables the laboratory to automatically add on and perform second line tests on patient samples, as clinically indicated based on the results of initial tests. [3] For example, an institution might offer a thyroid testing reflex protocol where laboratories first test TSH (thyroid stimulating hormone) and then add on additional second line testing needed for diagnosis [e.g., free thyroxine (free T4) and/or T3] as warranted based on the TSH results. Laboratories offering institutional reflex testing will generally offer clinicians a choice between a standalone test or the reflex protocol (e.g., clinician could choose TSH alone or TSH with reflex to T3 and free T4). Reflex testing solves several test utilization challenges. First, without reflex testing, clinicians might be inclined to order second line tests upfront, just in case they will be needed. This is likely to lead to test overutilization. Conversely, if clinicians do not order second line testing upfront, they will need to use manual processes to add on the needed second line test. In practice, testing cascades, where second line test selection requires knowing the results of first line tests, in the absence of reflex protocols, will at best unnecessarily consume laboratory and clinician resources and at worst will lead to oversights in which important second line tests do not get ordered or get delayed. Finally, in some cases, second line testing may involve esoteric biomarkers and clinicians may not know the optimal downstream tests to order. By automating second line test ordering, reflex testing removes the clinician burden and potentially error-prone process of esoteric test selection.

At present, reflex testing protocols typically involve one or several "if-then" rules to determine whether to add second line tests to patient specimens based on first line test results (e.g., *if* TSH is high *then* add Free T4). However, reliance on simple if-then rules, limits the scope of reflexing protocols to test ordering decisions that are reducible to simple rules. In practice, many test ordering decisions are nuanced and patient-specific [6] and would be difficult, if not impossible, to support using traditional reflex testing frameworks. This may explain, at least in part, why laboratories typically only offer a narrow menu of reflex testing options. Moreover, rule-based reflex protocols do not scale well since every testing scenario will generally require a custom, often institutionally developed set of rules. Improved reflex testing protocols, to the extent that they could appropriately encompass nuanced test ordering and be applied to widely varied situations, could provide a very valuable tool in supporting optimal test selection and ordering.

A potential solution to some limitations of the current standard of rule-based reflex protocols is to use machine learning to determine when to reflexively add on various tests, a concept we explore in this manuscript and call the "smart" reflex protocol. For example, machine learning based models could predict which second line tests are likely to benefit the patient and then automatically order these to be performed on existing specimens.

In this manuscript, we explore this concept of the "smart" reflex protocol, using ferritin, a commonly ordered blood-based biomarker of iron stores [7], as an exemplar. Low ferritin is virtually diagnostic of iron deficiency. While the vast majority of ferritin orders are for the diagnosis or monitoring of possible or confirmed iron deficiency, clinicians may also occasionally order ferritin to identify hemochromatosis, a genetic condition involving iron overload. Iron deficiency is most often suspected based on CBC test results; iron deficiency is a common cause of anemia, as identified by abnormally low hemoglobin and hematocrit parameters. Other CBC parameters including a low or low-normal MCV (or a decrease in MCV from a patient's baseline) and a high RDW may further suggest iron deficiency. However, no CBC finding in isolation can definitively diagnose or exclude iron deficiency, and no reflex testing protocols currently exist to add on ferritin studies based on the CBC results alone (in our lab and nor to the best of our knowledge any other clinical lab).

Here, we develop a proof-of-concept machine learning based smart reflex testing algorithm, designed to decide whether to reflexively add a ferritin test to a patient's CBC, using CBC results and other prior patient data. Our machine learning model is trained to predict whether a clinician would have manually ordered a second line ferritin test given the CBC results, using retrospective electronic medical record (EMR) data as the ground truth. That is, we assume that if a clinician manually ordered a ferritin on a patient after receiving the CBC results, then they would also have wanted to use a reflex protocol, were it deployed, to order ferritin on that patient. We validate our model's performance both against the retrospective ground truth and in comparison to simple if-then rules of the type that might be considered were a lab exploring traditional types of reflex protocols for ferritin.

2. **METHODS**

Our methods are summarized in Figure 1. As described in greater detail below, we extracted sets of clinical laboratory test results from patients who, at minimum, had complete blood count (CBC) testing. Using this set of laboratory test results along with other basic patient data, we developed a machine learning model to predict whether a patient would have a ferritin test ordered concurrent with and/or within 30 days of each CBC.

**Site, Dataset, Patient Inclusion and Ethics:**

Data for this study were derived from clinical laboratory testing performed at the Massachusetts General Hospital (MGH) and affiliated health centers. MGH is a large, tertiary care hospital based in Boston, MA and is a member of the Mass General Brigham (MGB) health system. All data used received institutional review board approval.

Data were extracted from the MGB clinical laboratory information system via an established laboratory data mart. Patients were included in this study if they had at least one outpatient CBC test during a one-year period ("study year") performed in the clinical laboratories at MGH. In addition to CBC test results, we extracted additional laboratory test results as shown in Table 1.

**Data Temporal Aggregation and Processing:**

The unit of analysis for our study was a "CBC-event" defined as an outpatient CBC during the study year. Each CBC was treated as a separate case for the purposes of our analysis. For each CBC-event, we defined and captured a dichotomous target variable ("ferritin ordered") denoting whether a ferritin was performed at the time of and/or within 30 days after the CBC. Additionally, we captured predictor variables including age, gender and laboratory testing data as summarized in Table 1. As shown, we included current CBC parameters as predictors, but no non-CBC lab results from the 30 days prior to the CBC. We chose not to include concurrent (and up to 30 days prior) results from non-CBC tests in part to avoid potential biases (for example that iron studies tend to be ordered together) and to facilitate the development of an algorithm that would perform well regardless of what additional labs (if any) beyond the CBC were ordered.   We additionally included predictor data across a range of analytes (summarized into defined features as described in Table 1), for test results occurring between two-years prior to the CBC up until 30 days before the CBC.  These data were aggregated temporally by taking the mean, count, minimum, maximum, and sum (equivalent to mean times count) of values observed during the historical period. All data were centered and scaled to have zero mean and unit variance.  Missing values were imputed to the nearest value for the recent CBC feature-set, and for the historical lab values any labs that were never measured were imputed to the population mean and the fact that no data were reported for that feature was captured in the count statistic. As a result of our technical approach to data extraction and aggregation, for cases in which a patient had multiple CBCs during the study year, prior results were considered up to two years prior to the first CBC during the study year; thus for a patient with a CBC both early in the study year and late in the study year, the prediction in reference to the later CBC would include predictor data as early as two years prior to the earlier CBC and thus up to three years before the index CBC event.

Of note, in very rare cases, individual CBC parameters may be ordered by the clinician without a full CBC being performed; for the purpose of this analysis, we assume the patient had a CBC if results were reported for any of the CBC parameters.  The sample collection date and time were used as the timestamp for all lab results.

**Data partitioning**

Data was randomly split into training, hyperparameter tuning, and testing partitions in an 80:10:10 ratio. Individual patients, even if enrolled multiple times due to multiple CBCs, were not split across partitions. Using a Monte-carlo cross validation framework, we split the data into 10 random such partitions, and repeated the same training, hyperparameter tuning, and validation procedure on each split, and results represent cross validation performance reported using the test sets for the optimal model configurations according to hyperparameter tuning in each split separately.

**Model Training and Validation**

We built logistic regression and random forest-based models using Scikit-learn [8]. Hyperparameter tuning was performed via a manual search over a variety of regularization parameters for both the Logistic Regression and Random Forest model types, and final results are based on whichever model type and hyperparameters maximized performance for the hyperparameter tuning partition for each partition. In all cases, random forest models outperformed logistic regression baselines, so final results use that model type, with mildly different regularization parameters per partition.

**Chart Review**

In a subset of cases where the model predictions disagreed with clinician decisions in the EMR, we additionally performed chart review. In particular, we identified the 200 cases where ferritin was ordered that had the lowest predicted probability of ferritin ordering and the 200 cases where ferritin was not ordered that had the highest predicted probability of ferritin ordering (probability cutoffs for each group are included below in the results sections). For the purposes of chart review, we considered cases that were included in the test partition for at least one of the first three (of the total ten) runs of the model and used the median predicted probability of ferritin ordering across any of the three runs for which the case fell into the test partition. From each set of 200 cases, we randomly selected 20 for chart review (40 cases total comprising 20 cases where ferritin was ordered and 20 where it was not). All cases selected for manual chart review were reviewed by one pathologist (JMB); half of the cases were also independently reviewed by a second pathologist (ASD) to capture inter-observer agreement. Pathologist review involved reviewing the electronic health record, including narrative notes and other findings to decide whether the patient *should* have been tested for ferritin (even if they were not actually tested in reality). The pathologist scored the cases on an ordinal scale as follows: *Ferritin Clinically Indicated*; *Ferritin Possibly Indicated*; *Ferritin Probably Not Indicated*; *Ferritin Not Indicated*. Considerations in the pathologist review included additional explanations for ordering or not ordering ferritin offered in the clinical notes and evidence of possible iron deficiency (or other reason to order ferritin) based on holistic review of relevant clinical notes and laboratory test results.

**Comparison to rule-based reflex testing protocols**

We compared model performance to a series of hypothetical rule-based triggers analogous to the type of rules that laboratories typically use in reflex testing protocols. Hypothetical rule sets used are shown in Table 2. Although to the best of our knowledge, no lab actually has a ferritin reflex testing protocol (perhaps due to the inadequacy of rule-based triggers in this context), we present this analysis to show why an algorithm would present a practical strategy even in cases such as this where rules fail.

In particular, we assessed the concordance between patients who would (and would not) have had ferritin ordered under each hypothetical set of reflex rules and the actual ferritin ordering in practice. In assessing predictive model performance in this sub-analysis, only test partition predicted probabilities were considered and for patient-CBCs included in the test partition on multiple training-test splits, the mean predicted probability across runs was used. The hypothetical rule based reflex protocols are shown in Table 2.

**Analytics and Statistics**

Modelling was performed in Python making use of Scikit-learn.[7] Outcome metrics were assessed using auROC, auPrC, and Brier loss, all calculated using Scikit-learn and aggregated over ten random train-test data splits. Analyses comparing the model to hypothetical reflex rules were performed in R [9] and plotted using the ggplot2 package [10] with confidence intervals for sensitivity and specificity calculated using the R binom package and Wilson method.

**RESULTS**

The final dataset included 288,427 unique "CBC-events" across 137,451 unique patients. 35,132 of the CBCs (12.2%) were associated with a ferritin within 30 days.

**Model Performance:**

Model performance is shown in Figure 2. As shown, the models provided moderate discriminative power in predicting whether ferritin would be ordered within 30 days as shown in the ROC curves (Figure 1A) and precision-recall curves (Figure 1B). The overall AUC was 0.731 (standard deviation=0.004) and AUPRC was 0.349 (standard deviation=0.011). The models were generally well-calibrated as shown in Figure 2B (Brier loss=0.094, standard deviation=0.003). Additional analysis using a modified definition for the ground truth labels is shown in Supplemental Appendix B and is highly consistent with our primary findings shown in Figure 2.

Figure 3 shows the features that were most important in model predictions. In general, these closely overlap with the clinical features that might suggest iron deficiency or that a physician might consider in deciding whether to order ferritin. Of note, in addition to clinical features, the model found whether the patient had a ferritin test within the prior two years (excluding the prior month) to be highly informative. This is in large part likely capturing the fact that patients with a history of iron deficiency or prior signs and symptoms requiring ferritin testing are more likely to undergo ferritin testing. It may also in part capture clinician style with respect to whether the patient is seeing a physician who tends to order ferritin.

**Comparison of machine learning to rule based reflex testing**

While to the best of our knowledge, no laboratory supports a ferritin reflex testing protocol (presumably due to the inadequacy of traditional rules for doing so) here we compared hypothetical rules involving hematocrit, MCV and RDW of the type that a lab might consider for reflex testing in other context to the performance of our machine learning model (Figure 4). As shown, most rule-based reflex testing protocols would be woefully inadequate while the model performs considerably better. For example, as illustrated in Figure 4, the most sensitive rule set tested (Low MCV or Low HCT) only achieved a 51% sensitivity; this rule had a specificity of only 63%. At that rule's sensitivity of 51%, the model could achieve a specificity of about 80%. Likewise, at that rule's specificity of 63%, the model could achieve a sensitivity of about 69%. While model performance would likely require further optimization (see discussion) for general clinical application, the current model may offer support for key applications as expounded in the discussion.

**Association between predicted probability of ferritin and ferritin result:**

As a secondary goal of this manuscript (alongside our primary goal of developing a proof of concept for smart reflex testing), we evaluated the association between model predicted likelihood of ferritin ordering and ferritin results as described in the supplement (Supplement A). We unsurprisingly find evidence (Supplemental Figure S1) that ferritin test results are not missing at random.

### Clinical Appropriateness of Predictions:

While we trained our models to predict whether ferritin would be ordered, an ideal reflex protocol would capture whether ferritin *should or should not* be tested. Thus, we performed chart review on selected cases (as described in the methods) where model predictions and actual test ordering were particularly discordant, representing 20 randomly selected cases out of the 200 where ferritin was ordered but the predicted probability of ordering was 4.7% or less and 20 cases of the 200 where ferritin was not ordered, but the predicted probability of order was 52% or greater. We found that in many cases the model predictions were more "correct" (i.e., appeared to be better in line with ideal clinical practice) than the actual clinical ordering (Table 3). Indeed, in multiple cases reviewed where the model predicted ferritin would be ordered but it was not, the failure to order ferritin appeared to be a simple oversight (e.g., note indicates intention to order ferritin and ultimate order includes other iron studies but not ferritin). Likewise, in many cases where ferritin was ordered in contradiction to model predictions, the ferritin appeared to be unnecessary. Inter-pathologist concordance in assessing the need for ferritin is shown in Table 4. In all but one case, the two pathologists arrived at the same conclusion or a conclusion within one level on the ordinal scale.

3. Discussion

Here we show that a machine learning model can predict whether a patient getting CBC testing will also get ferritin testing concurrently or within 30 days with a moderate degree of accuracy (Figure 2). Moreover, we show through chart review that in cases where the model prediction and actual ferritin test ordering are discordant, the model predictions may often be "correct" (and actual test ordering suboptimal) with respect to clinically optimal test ordering (Table 3). Indeed, in some cases where clinicians did not order a ferritin, but the model predicted they would, the failure to order ferritin appeared to be a clear oversight. Similarly, in some cases where ferritin was ordered in contrast to model predictions, the ferritin order appeared to have been placed by mistake, or at least without any clear clinical basis. Most importantly, we show that the model outperforms many simple rules of the type that drive traditional reflex testing protocols (Figure 4) and suggest, as expounded below, that this offers the potential for the model to serve as a foundation for smart reflex testing.

These findings suggest multiple potential clinical applications (Figure 5) to smart reflex testing. As shown in Figure 5, the general approach would be for a clinician to order a CBC with possible reflex to ferritin (and other iron studies). The algorithm would automatically add on ferritin as appropriate based on model computations performed after the CBC results become available. The performance of the current algorithm is likely inadequate for it to be used generally to fully replace manual ferritin test ordering (although we anticipate being able to improve performance through subsequent work as below). Nonetheless, our algorithm even with current performance could support important clinical scenarios (Figure 5). The first of these ("Variation 1"), would be the clinician who would order CBC plus iron studies upfront on a patient, anticipating the iron studies to be needed and not wanting to overlook them. In this case, the clinician could order a reflex protocol using a highly sensitive algorithm cutoff that when in doubt will perform ferritin. This algorithm could "cancel" the ferritin order when it can confidently predict it to be unneeded. A second application ("Variation 2") would be in patients where the clinician is simply ordering a CBC without a suspicion of iron. The algorithm could run at a highly specific cutoff helping to reduce the risk of overlooked iron deficiency. All these applications would require additional clinical validation.

Another advantage of our approach is potential generalizability. Unlike rule-based reflex testing that often requires each alert be set as a one-off, adaptations of the prediction models described here could potentially be trained across hundreds or thousands of test orders using a basic generalization of the approach developed herein. Of course, for many tests, it will be critical to pull non-laboratory clinical factors into the model for use as predictors.

While the primary aim of this paper was to explore smart reflex testing algorithms, we envision a secondary application of the algorithm we developed: reducing imputation bias when imputing missing laboratory test results in the development of machine learning models. A common framework for the development of machine learning based prediction models involves first imputing missing predictor data (e.g., missing laboratory test results) and then applying actual and imputed data to a downstream prediction model. However, most imputation methods assume missing data are missing at random (MAR) and may introduce biases if the MAR assumption is violated (as would typically be the case with missing lab results). [11-15] Indeed, our findings that the rate of abnormal ferritin results correlates with the likelihood of ferritin ordering (Supplemental Figure S1) suggests that the distribution of missing ferritin values differs from the distribution of measured values and in turn might suggest that typical imputation methods are likely to produce biased imputation results. We anticipate that use of models such as this for predicting whether a test will be ordered may help to correct for imputation bias. We aim to explore this in future work.

An important consideration is that clinicians use information, including clinical factors to decide whether to order ferritin, that go beyond the information available to our model. Thus, we envision being able to improve this model through future work by incorporating additional non-laboratory predictors.

4. **LIMITATIONS**

This manuscript is subject to several limitations. First, despite the inter-pathologist concordance, use of chart review for clinical validation remains subjective. However, chart review is only used however as a final feasibility assessment and many of our model evaluation metrics (e.g., those described in Figures 2-4) are not subjective. More importantly, for the approach developed in this manuscript to generate clinically optimal test ordering, the training data and hence current test ordering practice would need to represent clinically optimal test ordering. This consideration impacts not only our proposed smart reflex testing framework, but other intelligent decision support systems developed in the past[16]. In practice, actual test ordering is imperfect. Nonetheless, even if imperfect, the model may be sufficient to offer improvement on current practice, a hypothesis that may in part be supported by the chart review. Future solutions may also use hand-developed rules to filter out clearly non-optimal test orders (or missed orders) from the training data to improve clinically targeted performance. More rigorous validation studies would be needed prior to clinical implementation. Another limitation is that this study is based on a single center. Finally, we did not incorporate non-laboratory clinical predictors into our prediction models. We aim to explore doing so in future iterations and expect that this may improve performance.

5. **CONCLUSION**

This paper offers a framework to enhance clinical laboratory diagnostics through the development and deployment of machine learning-based "smart" reflex testing. We anticipate that models such as the one developed here have the potential to expand reflex testing across a wide range of patients and tests.

## 6. TABLES

**Table 1: Analytes Included as Predictors or Target Variations**

| Analyte Group | Analytes Included | Used as a target[1] | Concurrent results used as a predictor[2] | Aggregate historic results used as predictors[3] |
|---|---|---|---|---|
| **Ferritin** | Ferritin | ✓ | | ✓ |
| **CBC** | Hemoglobin, hematocrit, platelet count, MCV, MCH MCHC, RDW, RBC, WBC | | ✓ | ✓ |
| **Routine Chemistry** | Alanine transaminase; Albumin; Alkaline phosphatase; Anion gap; Aspartate transaminase; Bicarbonate; Blood urea nitrogen; Calcium; Chloride; Creatinine; Globulin; Glucose; Magnesium; Phosphorus; Potassium; Sodium; Total bilirubin; Total protein | | | ✓ |
| **Coagulation** | aPTT; D-dimer; INR; PT | | | ✓ |
| **Hematology** (non-routine CBC) | Absolute basophil count; Absolute eosinophil count; Absolute lymphocyte count; Absolute monocyte count; Absolute neutrophil count; Bands; Metas; Myelos; Percent basophils; Percent eosinophils; Percent lymphocytes; Percent monocytes; Percent neutrophils; Percent nucleated RBCs; Reactive lymphs; Retic; Schistocytes | | | ✓ |
| **Iron studies** (besides ferritin) | Iron; Total iron-binding capacity; | | | ✓ |
| **Lipids** | Cholesterol; HDL; LDL; Triglycerides | | | ✓ |
| **Other** | Amylase; B12; CRP; CEA; Creatine kinase; Direct bilirubin; ESR; Folic acid; Free T4; LDH; Lipase; NT-ProBNP; Osmolality; PTH; Lactic Acid; PSA; Testosterone; Troponin-T[4]; TSH; Uric acid; Urine total protein; Vitamin D | | | ✓ |

[1] Target variable was dichotomous describing whether a ferritin was performed concurrent with or within 30 days of the CBC
[2] The test results itself was used (or in rare cases where not reported on the current CBC, the most recent prior result within 30 days)
[3] The patients' the mean value of the analyte; the standard deviation of the values of the analyte on the patient; the number of times the analyte was tested on the patient("count"); the minimum and maximum values of the analyte; the sum of all results for the analyte (mean times count). All calculated across all of the patient's results from analyte collected from two years (in some cases as long as three years) prior up until one month prior to the CBC
[4] 4th generation troponin T, only; excluded more recent 5th generation results

**Table 2: Hypothetical Reflex Testing Rules Used for Comparison to the Model**

| Rule | Definition |
| --- | --- |
| Low HCT | Patient's HCT is below the reference limit |
| Low MCV | Patient's MCV is below the reference limit |
| Prior Ferritin | Patient had ferritin test in the two years prior to the CBC |
| Prior Ferritin and Low HCT | The patient had ferritin test performed in the two years prior to the CBC AND the patient's current HCT is below the reference limit |
| Low MCV and High RDW | The patient's MCV is below the reference limit AND the RDW is above the reference limit |
| Low HCT or MCV | The patient's HCT is below the reference OR the patient's MCV is below the reference or both |
| Prior Ferritin and Low MCV | The patient had ferritin test performed in the two years prior to the CBC AND the patient's current MCV is below the reference limit |
| Low HCT and MCV, High RDW | The patient's HCT and MCV are both below their respective reference limits and the RDW is above the reference limit |
| **These are representative of the types of rules a laboratory might use to develop a reflex protocol for ferritin. In practice, we are not aware of any labs with such a ferritin reflex protocol, perhaps due to the inadequacy of these rules (and hence the motivation for this paper).** {colspan=2} | |
| MCV = mean cell volume; HCT = hematocrit; RWD = red cell distribution width. {colspan=2} | |
| Gender specific, adult reference ranges were used (regardless of actual patient age); HCT: male 41%, female 36%; MCV 80 fL; RDW 14.5%. {colspan=2} | |

**Table 3: Chart Review Findings**

| Pathologist Assessment | Model Prediction Ferritin tested[1] | Ground truth Ferritin **not** tested | Model Prediction Ferritin **not** tested[2] | Ground truth: Ferritin tested |
|---|---|---|---|---|
| Ferritin Clinically Indicated | 10 | | 1 | |
| Ferritin Possibly Indicated | 2 | | 4 | |
| Ferritin Probably Not Indicated | 3 | | 4 | |
| Ferritin Not Indicated | 5 | | 11 | |

[1] 20 randomly selected cases from the 200 with the greatest predicted probability of ferritin testing where ferritin was not tested
[2] 20 randomly selected cases from the 200 with the lowest predicted probability of ferritin ordering where ferritin was tested

Shown are the findings of a pathologist in reviewing cases. 20 of the 200 test cases where the model most confidently predicted ferritin would be tested as well and 20 of the 200 test cases where the model most confidently predicted ferritin would not be tested. Cells represent the number of cases.

**Table 4:** Pathologist Chart Review Concordance

| | | Pathologist 2 | | | |
|---|---|---|---|---|---|
| | | Ferritin Clinically Indicated | Ferritin Possibly Indicated | Ferritin Probably Not Indicated | Ferritin Not Indicated |
| **Pathologist 1** | **Ferritin Clinically Indicated** | 5 | 0 | 0 | 0 |
| | **Ferritin Possibly Indicated** | 2 | 0 | 1 | 0 |
| | **Ferritin Probably Not Indicated** | 1 | 1 | 0 | 1 |
| | **Ferritin Not Indicated** | 0 | 0 | 0 | 10 |

Shown is the number of cases based on the assessments of the two pathologists. Numbers shown indicate the number of cases. Pathologist 1 (JMB) is the primary pathologist whose findings are shown in Table 3. Green cells represent complete agreement while yellow cells represent agreement within one level on our ordinal scale. Only a subset of the cases shown in Table 3 were reviewed by both pathologists and shown here.

## 7. FIGURE LEGENDS

### Figure 1: Methods Overview

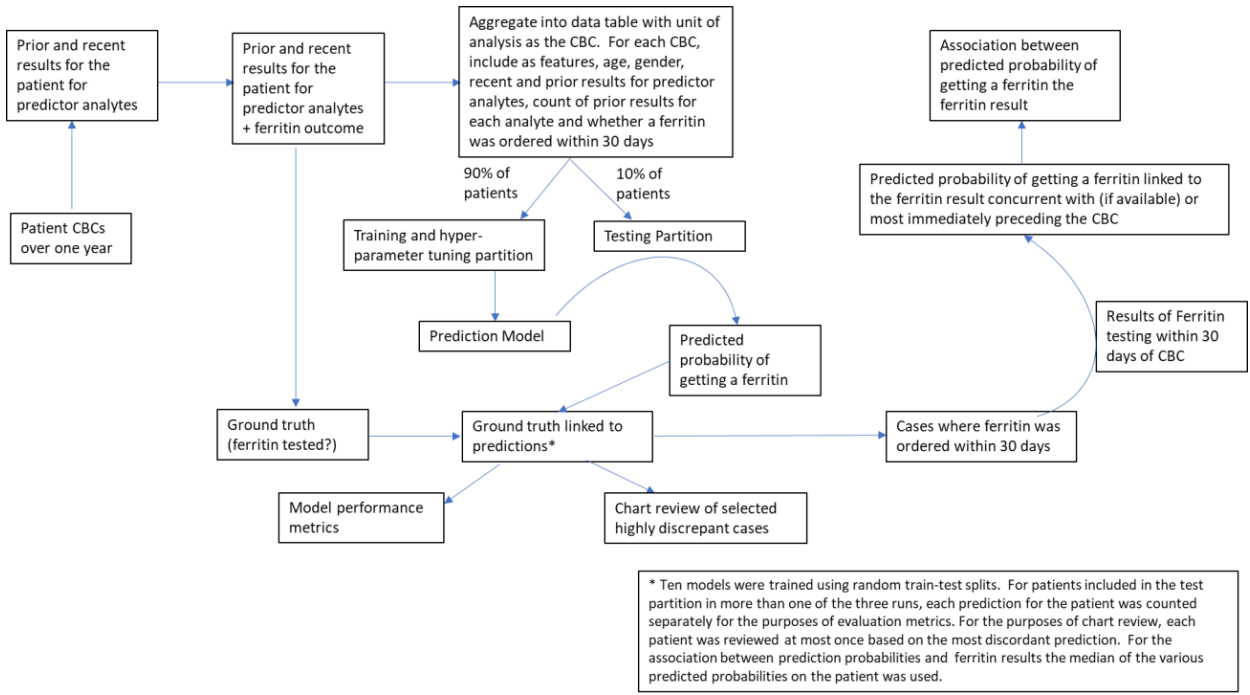

Shown is a flowchart providing an overview of the dataflow and methods we used to develop and test our prediction model.

**Figure 2: Model Performance**

Shown is a calibration curve (A), a receiver Operating Characteristic Curve (B) and precision recall curve (C) for the test partition aggregated across the ten runs. Ribbons represent mean plus/minus standard deviation across all 10 runs.

A.

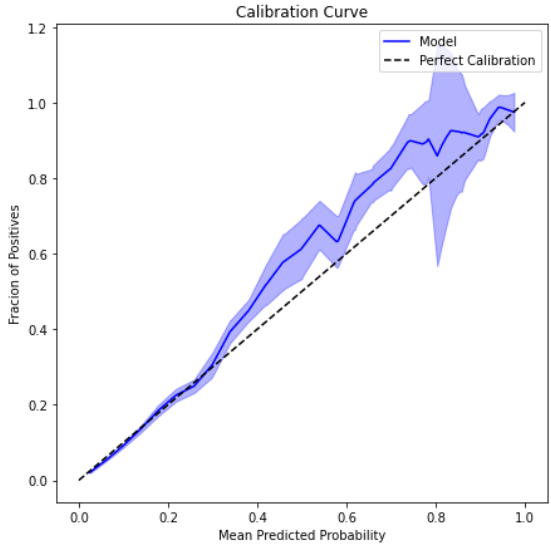

B.

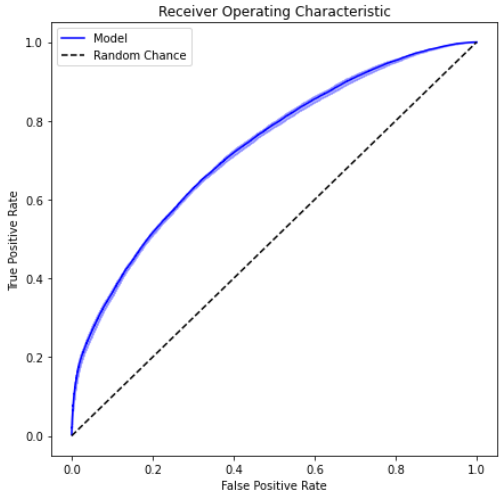

C.

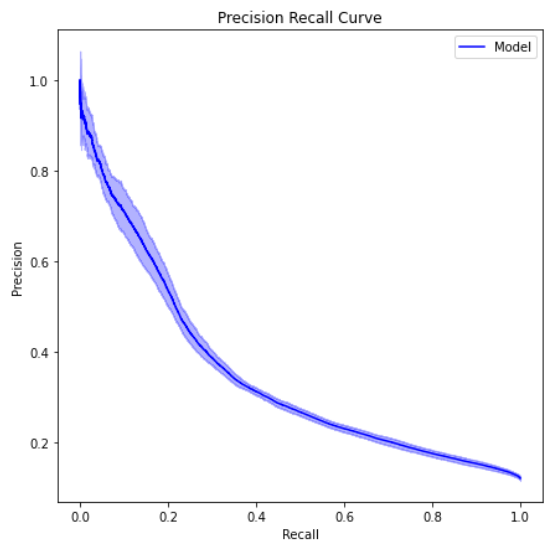

**Figure 3: Feature Importance**

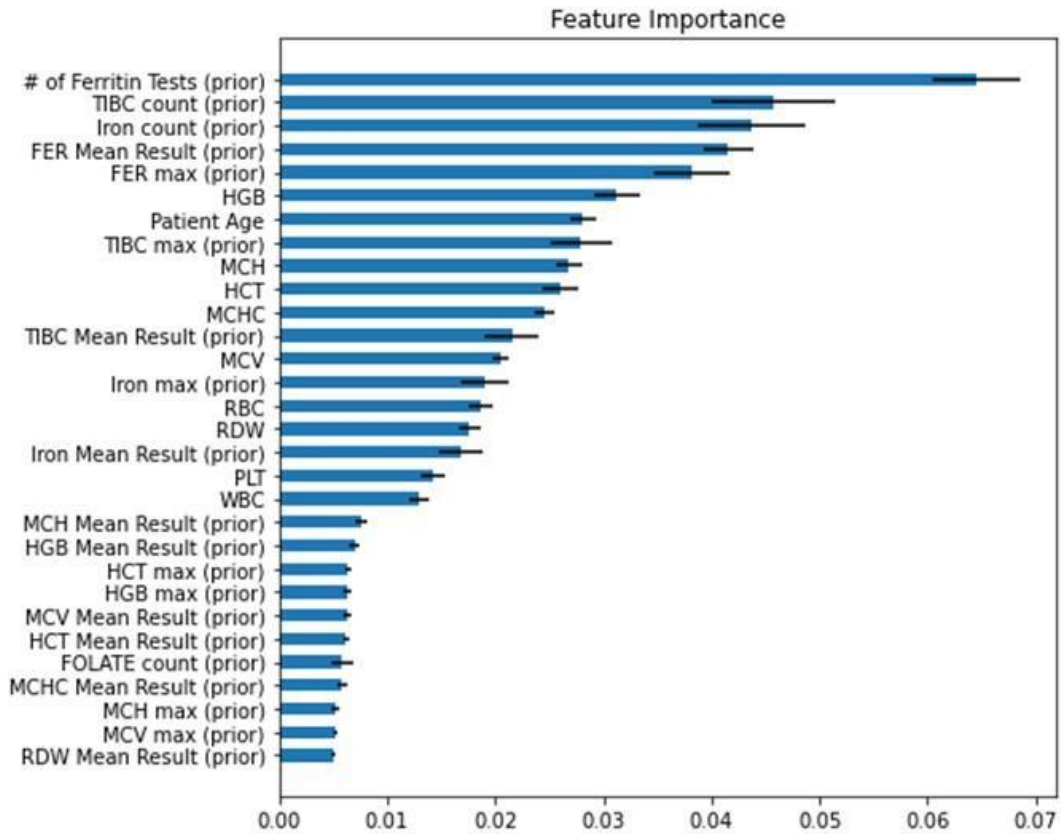

Aggregated feature importance (higher is more important) across all models (mean plus/minus standard deviation), as judged by the average decrease in the gini entropy when this feature is used across all decision trees in the random forest. "(prior)" indicates the historical value of the feature (older than 30 days ago) is being referenced, rather than the current results for that test.

**Figure 4: Comparison to hypothetical rule-based reflex testing protocols**

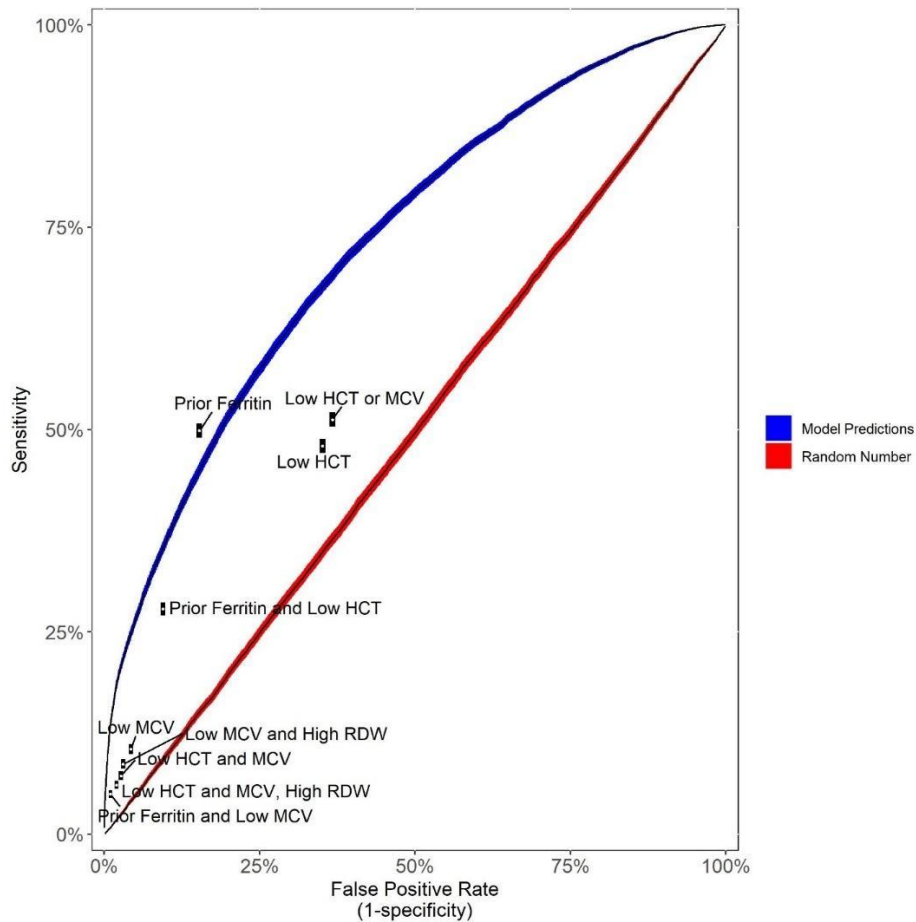

Shown is an ROC-like plot describing the concordance between actual ferritin test ordering and both i) the predictive model (blue curve; looks across various probability thresholds in the test data) and ii) various hypothetical rule-based testing triggers (labeled white points). Performance of randomly adding on ferritin (red line) is also shown as a negative control. False positive rate (x-axis) represents the proportion of patients who did not have ferritin in actual practice who would have using the model (at a given cutoff) using the specified hypothetical reflex testing rules. Sensitivity (y-axis) represents the proportion of patients who had ferritin in actual practice who would have using the model (at a given cutoff) or specified rules. Line width (blue and red curves) represent 95% confidence intervals. For hypothetical rules, white points represent point-estimates and black boxes surrounding white points represent 95% confidence intervals. By way of example, the most sensitive rule shown ("Low HCT or MCV") has a sensitivity and specificity of only 51% and 63%. The model could offer a sensitivity and specificity of 51% and 80% or of 63% and 69% (among other combinations shown in the blue curve). (HCT=hematocrit, MCV= mean cell volume, RDW=red cell distribution width).

**Figure 5: Clinical Applications to Smart Reflex Testing**

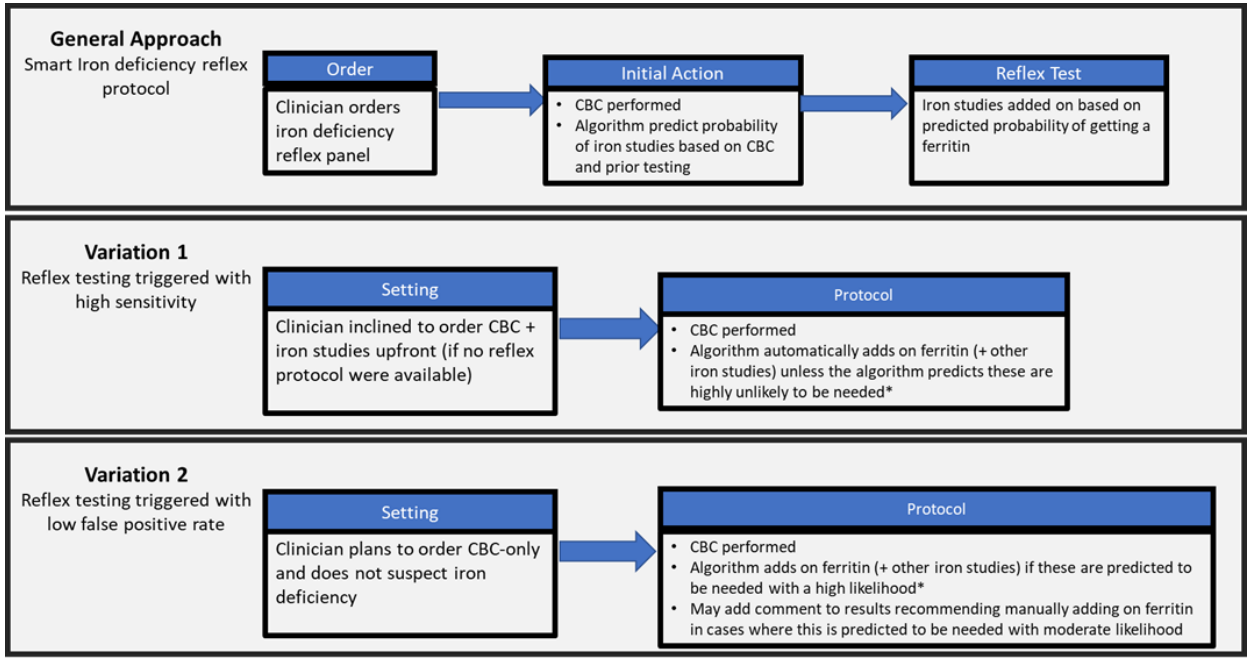

Potential applications to smart reflex testing are illustrated. The general case would require further model development and variations 1 and 2 would require further validation prior to implementation in clinical practice.

8. **FUNDING STATEMENT**

    This work was support by National Library Medicine grant R01 LM013337.

9. **COMPETING INTERESTS STATEMENT**

    Dr. Baron is an employee of Roche Diagnostics in addition to his academic role.

10. **CONTRIBUTOR STATEMENT**

    Project framework and design:  All Authors
    Model development:  MM
    Data structuring and analysis and model evaluation:  MM (primary); JB (secondary)
    Chart review:  JB and ASD
    Manuscript initial drafting: JB and MM
    Manuscript revision:  All authors

11. **Data Sharing**:  Raw patient data cannot be shared due to privacy constraints and institutional policies.

**Supplement A: Association between ferritin results and predicted probabilities**

*Methods:* For cases where ferritin testing was performed concurrent with or within 30 days after the CBC, we assessed the ferritin test results to look for an association between predicted probability of ferritin ordering and results. We grouped cases by decile based on their test partition predicted probability of getting a ferritin. For cases that were in the test partition on multiple runs, the median predicted probability was used. For each decile, we calculated the proportion of ferritin results that were abnormally low (below our laboratory's reference cutoffs of 10mg/dL for females and 30mg/dL for males). This portion of the analysis was performed in R [9] with confidence intervals calculated using the Wilson method for binomial confidence intervals as implemented in the R binom package. Results were plotted in R using the ggplot2 package [10].

*Results:*

As shown (Figure S1), patients in whom the model predicted a particularly low likelihood of having a ferritin ordered (e.g., decile 1 in Figure 4) very rarely had abnormally low ferritin results. Presumably in most ferritin orders in patients with predicted likelihoods in the first decile were unnecessary and would have been avoided using an optimized reflex testing protocol.

Likewise, this demonstrates that ferritin test results are not missing at random and thus methods to impute ferritin results in the context of predictive model development are likely to be biased.

**Figure S1: Association Between Probability of Getting A Ferritin And The Ferritin Result**

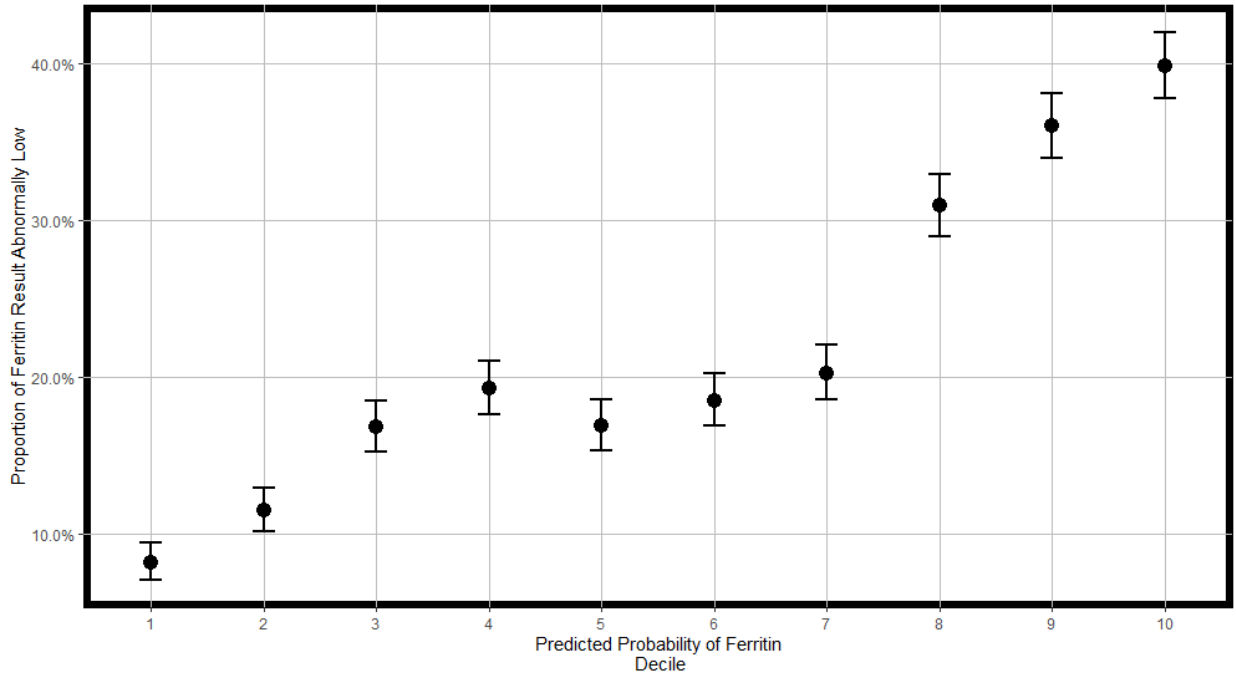

Shown is the relationship between predicted probability of ferritin (x-axis) and the probability that ferritin will be abnormally low (y-axis) for cases where ferritin was actually ordered. Error bars represent 95% confidence intervals. Spearman rho assessing the correlation between predicted probability of ferritin (by decile) and proportion of abnormal results was 0.96.

**Supplement B: Refinement of Ground Truth Labels**

After completing our primary work, we discovered that in 541 cases (~0.2%) the ground truth ferritin-ordered labels, while literally correct per the criteria defined in the methods section, were arguably misleading. In particular, these cases were labeled as not having a ferritin within 30 days after the CBC, but the patient indeed had a ferritin shortly *before* (usually 1-3 minutes) the CBC, presumably artifactually related to tube label print times. Given the small number of cases affected, we expected the impact to be minimum. Nonetheless out of an abundance of caution, we re-ran the analysis labeling cases where the ferritin was collected up to 1 hour prior to the CBC as having a ferritin test concurrent or within 30 days. As expected, results were not materially changed (updated auROC = 0.729±0.004, auPRC= 0.349±0.01, Brier Loss: 0.095±0.003).